\documentclass[10pt,twocolumn,letterpaper]{article}
\usepackage{wacv}
\usepackage{times}
\usepackage{graphicx}
\usepackage{amsmath}
\usepackage{amssymb}

\usepackage{pseudocode}
\usepackage{multirow}
\usepackage{array}
\usepackage{times}
\usepackage{enumerate}
\usepackage{balance}

\wacvfinalcopy 

\def\wacvPaperID{131} 

\ifwacvfinal\fi

\begin{document}


\title{Summarisation of Short-Term and Long-Term Videos using Texture and Colour}

\author{\it Johanna Carvajal, Chris McCool, Conrad Sanderson\\
~\\
{NICTA, GPO Box 2434, Brisbane, QLD 4001, Australia~\thanks{{\bf Acknowledgements:} NICTA is funded by the Australian Government through the Department of Communications and the Australian Research Council through the ICT Centre of Excellence Program.}}\\
{University of Queensland, School of ITEE, St Lucia, QLD 4072, Australia}\\
{Queensland University of Technology (QUT), Brisbane, QLD 4000, Australia}
}

\maketitle
\thispagestyle{empty}

\begin{abstract}

\noindent
We present a novel approach to video summarisation that makes use of a Bag-of-visual-Textures (BoT) approach.
Two systems are proposed, one based solely on the BoT approach and another which exploits both colour information and BoT features.
On 50 short-term videos from the Open Video Project we show that our BoT and fusion systems both achieve state-of-the-art performance, obtaining an average F-measure of 0.83 and 0.86 respectively,
a relative improvement of 9\% and~13\% when compared to the previous state-of-the-art.
When applied to a new underwater surveillance dataset containing 33 long-term videos, 
the proposed system reduces the amount of footage by a factor of 27, with only minor degradation in the information content.
This order of magnitude reduction in video data represents significant savings in terms of time and potential labour cost when manually reviewing such footage.
\end{abstract}

\section{Introduction}

Video abstraction aims at providing concise representations of long videos.
It has applications in browsing and retrieval of large volumes of videos~\cite{Ajmal2012} and also in improving the effectiveness and efficiency of video storage~\cite{Truong2007}.
Video abstraction can be categorised into two general groups: video summarisation and video skimming~\cite{Qing-Ge2013,Truong2007}. 
Video summarisation, also known as still image abstraction, static storyboard or static video abstract, is a compilation of representative frames selected from the original video~\cite{deAvila2011}.
Video skimming, also known as moving image abstraction or moving/dynamic storyboard, is a collection of short video clips~\cite{Almeida2012,Qing-Ge2013}. 
Both approaches should preserve the most important content from the video in order to present a comprehensible and understandable description for the end user. 

In general, video skimming provides a more coherent and visually attractive result.
It often retains a high-level of linguistic meaning due to its capacity to combine audio and moving elements~\cite{JungHwan2004,Truong2007}. 
However, video summarisation is easier to generate and is not constrained in terms of timing and synchronisation~\cite{Almeida2012,Truong2007}.

Video summarisation is an active area of research within the computer vision community and it has been applied in various video categories such as  Wildlife Videos~\cite{Suet-Peng2012}, sports videos~\cite{Ouyang2013}, TV documentaries~\cite{Almeida2012}, among others.
In~\cite{Ajmal2012} the various approaches to video summarisation are divided into six techniques consisting of: feature selection, clustering algorithms, event detection methods, shot selection, trajectory analysis and the use of mosaics.
Often a combination of techniques is used, for example one of the most common approaches is to combine feature selection with a form of clustering~\cite{Almeida2012,deAvila2011,Mundur2006}. 

In~\cite{Zhuang98} a video summary is obtained by extracting a feature vector from each frame and then clustering the resulting set of feature vectors.
The smallest clusters are then removed.
A keyframe -- a frame that forms part of the video summary -- is selected for each cluster centroid by taking the frame whose feature vector is closest to the centroid.
Similar approaches are adopted in~\cite{Almeida2012,deAvila2011,Divakaran2001,Qing-Ge2013} where the major difference is in the choice of feature vector used to represent each frame.
Colour histograms are used in~\cite{Almeida2012,deAvila2011}, motion-based features are used in~\cite{Divakaran2001}, and saliency maps are used in~\cite{Qing-Ge2013}.
Each of the previously proposed feature vectors has its drawbacks.
For instance, the colour histogram approach used in~\cite{Almeida2012,deAvila2011} retains only coarse information about the frame.
Motion-based features of~\cite{Divakaran2001} fail when the motion in the videos is too large.
Finally, the saliency maps used in~\cite{Qing-Ge2013} perform poorly for cluttered and textured backgrounds.
To date, limited work has been done on incorporating texture information to perform video summarisation.
 
\textbf{Contributions.} In this paper we first propose the use of texture information to improve video summarisation.
We propose the use of the computationally efficient and effective bag-of-textures approach;
we conjecture that this will improve video summarisation as it has been successfully applied to a range of image processing tasks,
such as matching and classification of natural scenes and faces~\cite{Lin2010,Conrad2009,Jianchao2009}. 
The bag-of-textures model divides an image into small patches, extracts appearance descriptors from each patch, quantises each descriptor into a discrete ``visual word'',
and then computes a compact histogram representation~\cite{Grauman2011}, providing considerably different information than colour histograms.
In addition, we propose a fusion based system for video summarisation, where both colour and texture information is exploited.
This will allow us to overcome the shortcomings of either approach.
Similar approaches have been shown to be advantageous in object classification tasks~\cite{Li10_1:conference}.
We show that our system may be applied not only to short-term videos but also to long-term videos, helping in the detection of the existence of a rare species of fish.

The layout of this paper is as follows. 
In Section \ref{sec:VIS} we describe in detail our proposed video summarisation method that exploits the benefits of using texture histograms based on the bag-of-textures model. 
In Section \ref{sec:FVIS} we present our improved video summarisation method that fuses the visual information provided by both the colour and texture histograms. 
In Section \ref{sec:eva} we describe how we evaluate the video summaries of short-term and long-term videos. 
In Section \ref{sec:experiments}, we present experiments which show that the proposed methods obtain higher performance than existing methods based on colour histograms. 
Section \ref{sec:conclusion} summarises the main findings.

\section{Bag-of-Textures for Video Summarisation}
\label{sec:VIS}
 
This section describes our proposed bag-of-textures (BoT) approach.
There are four main stages:

\begin{enumerate}
\setlength{\itemsep}{0ex}

\item Pre-processing: The input video is sub-sampled after which each frame is filtered and rescaled.

  \item BoT representation:

  \begin{enumerate}[(i)]
  \setlength{\itemsep}{0.25ex}
    \item \textit{Local Texture Features}. Each frame is divided into small patches (blocks) and from each block we extract 2D-DCT features, which is an effective and compact representation~\cite{Pennebaker93_1:book}.
    \item \textit{Dictionary Training}. A generic visual dictionary is trained to describe the most commonly occurring textures in an independent training set. 
    \item \textit{Generation of BoT Histogram}. Each frame is represented by a histogram which is obtained by matching the feature vectors from each block to the dictionary. 
  \end{enumerate}

  \item Keyframe selection: Similar frames are grouped into an automatically determined number of clusters. One keyframe is selected per cluster.  
  \item Post-processing: In this final stage, we eliminate possible repetitive frames and create the static video summary.
\end{enumerate} 

\noindent Each of these stages is elucidated in the following sections.

\subsection{Pre-processing}\label{sec:pre}

\subsubsection{Sampling and Rescaling} The original input video is re-sampled to one frame per second in order to reduce the number of video frames to be examined.
Each frame is then converted into gray-scale and re-scaled to be a quarter of its original size, in order to reduce the computational cost of the following stages.

\subsubsection{Noise Filtering} There are often uninformative frames that appear at the beginning and/or the end of a segment that may affect the appearance of a video summary~\cite{deAvila2011}.
These frames are usually colour-homogeneous due to fade-in and fade-out effects, and have a small standard deviation of their pixel values.
Frames with a standard deviation below a threshold are eliminated.

\subsection{BoT Representation}\label{sec:fex}

\subsubsection{Local Texture Features}\label{ssec:dct}

Each frame is divided into $N$ overlapping blocks.
To each block we apply the 2D discrete cosine transform (2D-DCT) to obtain a $D$-dimensional feature vector that represents the local texture information~\cite{Pennebaker93_1:book}.
Thus, the local texture feature for the $n$-th block of the $i$-th frame is ${\bf{x}}_{i,n}$.

\subsubsection{Dictionary Training} \label{ssec:bof}

The dictionary is trained using the {\it k}-means algorithm~\cite{Bishop2006} by pooling the local texture features from a set of training frames.
The resulting $G$ cluster centers $\{{\boldsymbol{\mu}}_1,  \cdots, {\boldsymbol{\mu}}_{G}\}$ represent the local textures (codewords) of the dictionary.

\subsubsection{Generation of BoT Histogram}\label{ssec:fext} 

In the BoT approach the $i$-th frame is represented by a histogram, ${\bf {h}}^\text{BoT}_{i}$.
This $G$-dimensional histogram represents the relative frequency of the local texture features within the frame.
The $g$-th dimension of ${\bf {h}}^\text{BoT}_{i}$ is the relative frequency of the $g$-th local texture feature from the dictionary, similar to \cite{Dardas2010}.
The histogram is normalised to sum to one.
Thus, each local texture feature can be converted to a local histogram, ${\bf{h}}^{\text{BoT}}_{i,n}$, of dimension $G$ where each dimension $g$ is given by,
\begin{equation}\label{eq:h_in}
	{h}^{\text{BoT}}_{g,i,n} = \begin{cases} 1 & \text{if }g = \underset{{k \in 1, \cdots, G}}{\arg\ \min} \Vert {\bf{x}}_{i,n} - {\boldsymbol{\mu}}_k \Vert_{2} \\ 0 & \text{otherwise} \end{cases}.
\end{equation}
\noindent These $N$ local histograms can then be summed and normalised to produce the final BoT histogram,
\begin{equation}
	\label{eq:ave_h}
	{\bf {h}}^{\text{BoT}}_{i} = \frac{1}{N} \sum\nolimits_{n=1}^{N}{\bf{h}}^{\text{BoT}}_{i,n}.
\end{equation}

\subsection{Keyframe Selection}\label{sec:KmeansBoF}

To obtain a set of keyframes we adopt an approach similar to that of~\cite{deAvila2011}.
A keyframe is a frame that forms part of the video summarisation.
The {\it k}-means algorithm is used to cluster similar frames into $K$ segments, and the resultant centroids are then used to select the keyframes.

Initially, the frames are grouped consecutively, assuming that sequential frames share similar content.
To automatically determine the number of clusters, $K$, we calculate the Euclidean distance between two consecutive frames. 
If the distance is greater than a threshold $\tau$ then $K$ is incremented.
For each cluster centroid the frame whose BoT histogram is closest is selected as a keyframe.
A total of $K$ keyframes is then reached.

\subsection{Post-processing}\label{sec:post}

Having obtained the initial set of $K$ keyframes we then attempt to discard those keyframes which are too similar.
This is achieved by comparing all keyframes against each other.
If the Euclidean distance between the BoT histograms of the keyframes is smaller than a threshold $\tau$ then one of the two keyframes under consideration is discarded. 
This gives the final static video summary that consists of $N_{as}$ keyframes, where $N_{as}\leq K$,
with $as$ standing for \underline{a}utomatic \underline{s}ummary.

Lastly, the static video summary is obtained after organising the resulting keyframes in temporal order.

\section{Fusion of Colour and BoT}
\label{sec:FVIS} 

In this section, we present a hybrid system that fuses colour histograms~\cite{deAvila2011} and BoT texture information, termed as CaT (for \textbf{C}olour \textbf{a}nd \textbf{T}exture).
The proposed CaT approach to video summarisation has the same 4 stages as our proposed BoT video summarisation approach, but with additions in order to obtain colour histograms.
We describe these additions below.

\begin{enumerate}
\setlength{\itemsep}{0ex}

\item
Pre-processing: The input video is processed in two independent ways. 
First, we obtain the BoT histograms as described in Section~\ref{sec:pre}. 
Second, to obtain the colour histograms we extract the Hue component, from the HSV colour space, of the unscaled input frame similar to~\cite{deAvila2011}.
In both cases we remove uninformative frames by employing the noise filtering process described in Section~\ref{sec:pre}.

\item Texture and Colour Histogram: The BoT histogram is the same as explained in Section \ref{sec:fex}. 
The colour histogram, ${\bf{h}}^\text{hue}_{i}$, of the $i$-th frame is computed using only the Hue component as in~\cite{deAvila2011}.

\item
Keyframe Selection: 
The BoT and colour histograms are clustered using $k$-means.
This stage is similar to Section~\ref{sec:KmeansBoF}.
The difference lies in the distance measure used to compare all frames against each other. 

\begin{enumerate}[(i)]
\setlength{\itemsep}{0ex}

  \item To select the number of keyframes $K$ we combine the information from the BoT and colour histograms.
When calculating the distance between frame $a$ and $b$ we use the weighted summation of Euclidean distances:
\begin{equation}
  \label{eq:weighted_euclid}
  \alpha \Vert {\bf{h}}^{\text{BoT}}_{a} -  {\bf{h}}^{\text{BoT}}_{b}  \Vert_{2} + \beta \Vert {\bf{h}}^{\text{hue}}_{a} - {\bf{h}}^{\text{hue}}_{b}   \Vert_{2}
\end{equation}

\noindent
under the constraints $\alpha+\beta=1$, $\alpha \geq 0$, $\beta \geq 0$.

\item
Each keyframe is selected by finding the frame which is closest to each cluster centroid.
For the CaT approach the distance between a frame and a centroid is calculated as a weighted summation of the Euclidean distances, as per~\eqref{eq:weighted_euclid}.

\end{enumerate}

\item Post-processing: To eliminate similar frames we use the procedure described in Section~\ref{sec:post} but replace the Euclidean distance with the weighted summation of the Euclidean distances, as per~\eqref{eq:weighted_euclid}.

\end{enumerate}

\section{Datasets and Evaluation Metrics}
\label{sec:eva}
To evaluate the performance of video summarisation we use two datasets consisting of short- and long-term video data.
The short-term data is obtained from the Open Video Project\footnote{Open Video Project: {\it http://www.open-video.org}}.
The long-term data is a new dataset that consists of 14 hours of underwater video surveillance which monitors the behaviour of marine wildlife.

\subsection{Short-Term Videos}

We use the 50 videos from the Open Video Project which contain ground truth~\cite{deAvila2011}.
Each ground truth consists of the summary provided by $P=5$ users.
The users provided the summaries under no restrictions upon length nor appearance of the summaries. 

To evaluate the performance on the short-term video data we use the ``Comparison of User Summaries'' (CUS) method~\cite{deAvila2011}.
This method compares the automatic video summarisation and ground truth by exhaustively calculating the distance between the frames from the automatic summarisation and the ground truth.
Two frames are similar if the distance between their respective feature vectors (histograms) is less than an evaluation threshold $\delta$.
If the frames match they are removed from the next iteration of the comparison process. 
For performance evaluation, the distance measure used for the BoT approach is the Euclidean distance, however, to be consistent with prior work~\cite{deAvila2011},
the distance measure for the colour histograms is the $L_{1}$-norm.
Therefore, the distance measure used for CaT is the weighted summation of the Euclidean distance for the BoT histograms and the $L_{1}$-norm for the colour histograms:
\begin{equation}
  \alpha \Vert {\bf{h}}^{\text{bof}}_{a} -  {\bf{h}}^{\text{bof}}_{b}  \Vert_{2} + \beta \Vert {\bf{h}}^{\text{hue}}_{a} - {\bf{h}}^{\text{hue}}_{b}   \Vert_{1}.
\end{equation}

Various evaluation metrics exist to measure the quality of an automatic video summary.
We use three evaluation metrics so that we can compare our proposed approaches with two state-of-the-art methods~\cite{deAvila2011,Almeida2012}.
To compare with~\cite{deAvila2011} we use accuracy ($acc$) and error ($err$), and to compare with~\cite{Almeida2012} we use the $F$-measure.

To calculate $acc$ and $err$, each frame in the automatic video summary is compared with all frames in the user summary and then the number of matching frames ($N_{m}$) and non-matching frames ($N_{nm}$) are calculated:
\begin{equation}\label{eq:eva_vsumm}
  \begin{array}{cc}
    acc = \frac{N_{m}}{N_{u}}, & err = \frac{N_{nm}}{N_{u}}  
  \end{array}
\end{equation}
\noindent where $N_{as}$ and $N_{u}$ are the total number of frames from the automatic and user summary, respectively.

The $F$-measure, defined as 
\begin{equation}\label{eq:eva_vison}
    F= \frac{2\times \text{precision}\times \text{recall}}{\text{precision} ~+~ \text{recall}}
\end{equation}
\noindent is used to to provide a single number that balances $\text{precision} = N_m/N_{as}$ and $\text{recall} = N_m/N_{u}$.

The evaluation metrics are presented as an average.
First, we take the average from the $P$ users to obtain $acc_P$, $err_P$, and $F_P$; for each video there are $P=5$ users.
Then we take the average across all of the videos to obtain $\overline{acc}$, $\overline{err}$, and~$\overline{F}$. 
In terms of $\overline{acc}$ it is desirable to have a high value as it measures the number of matching frames.
In terms of $\overline{err}$ it is desirable to have a small value as it measures the number of non-matching frames. 
With regards to $\overline{F}$ it is desirable to obtain a high value, which occurs when the $\text{precision}$ and $\text{recall}$ are large.

\subsection{Long-Term Videos}
\label{sec:eval:long_term}

\begin{figure}[!tb]
  \centering
  \includegraphics[width=0.8\columnwidth]{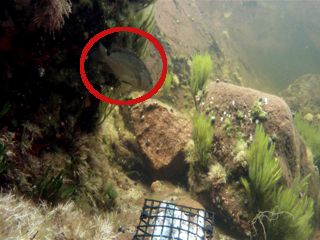}\\
  \vspace{1ex}
  \includegraphics[width=0.8\columnwidth]{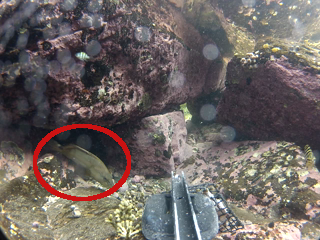}
\caption
  {
  Example images from the long-term underwater surveillance videos; the added red ellipsoids highlight the rare species of interest.
  }
\label{fig:bcod_images}
\vspace{-2ex}
\end{figure}

The long-term videos consist of 14 hours of underwater footage from 33 videos which are on average 25 minutes in duration.
This data was obtained from the NSW-DPI\footnote{New South Wales Department of Primary Industries, Australia.},
courtesy of David Harasti.
Example images are shown in Figure~\ref{fig:bcod_images}.
In each video there is always at least one segment where a rare species of fish, the black cod, is within view.
Normally these videos would be inspected by a human expert to determine if there is an instance of the rare fish within.
We propose that video summarisation can be used to reduce the amount of footage to be viewed in order to detect the existence of this rare species of fish.

Using ground truth which provides time-stamps when this rare species is within view, we examine the effectiveness of video summarisation to provide at least one keyframe in each static video summary with the rare species of interest within view.
This is useful as it presents a way to reduce the time and cost of manually viewing a large amount of video data.

To calculate the performance of long-term videos we present results in terms of detection accuracy and the average compression ratio ($R_{\text{c}}$).
Detection accuracy refers to whether an instance of the rare species is among any of the chosen keyframes for a static video summary; $75\%$ would mean that there is at least 1 keyframe of the rare species in $75\%$ of the static video summaries.

To calculate the average compression ratio we first note that because we have long-term videos then for each video there might be many hundreds of keyframes.
To present all of these keyframes effectively to the user we re-encode them into a static video summary by presenting each keyframe for $0.25$ seconds.
This gives the user time to effectively view the keyframe.
Thus the $t$-th long-term video ${\bf{V}}_{t}$ is converted to a static video summary ${\bf{S}}_{t}$ with a compression ratio given by:

\noindent
\begin{equation}
  R_{c,t} = 4 \times \frac{\text{Duration} ({\bf{V}}_{t}) } {\text{Duration} ({\bf{S}}_{t}) }
\end{equation}

\noindent
where $\text{Duration}$ is the duration of a video and the factor of $4$ is introduced as there are $4$ keyframes per second of the shortened video.

\section{Experiments}
\label{sec:experiments}

An important part of both the BoT and CaT approaches is the training of the dictionary to obtain the texture histograms.
To train this dictionary we use 10 frames randomly selected from videos taken from the Open Video Project that have no user summaries, ensuring they are independent of the evaluation dataset.
In addition, the frames selected to train the dictionary look significantly different to the ground truth provided by the users.

To obtain the proposed local texture features we divide each frame into a set of overlapping blocks.
Similar to~\cite{Conrad2009} we use a block size of $8 \times 8$ with an overlap margin of 6 pixels,
and represent each block as a $D=15$ dimensional feature vector containing 2D-DCT coefficients.
We extract the first 16 2D-DCT coefficients, which represent low-frequency information~\cite{Pennebaker93_1:book},
and omit the first coefficient as it is the most sensitive to illumination changes.
With regards to the colour histogram, we quantise the Hue component into 16 bins as per~\cite{deAvila2011}.
These parameters are the same for all experiments.

The values for the threshold $\tau$, fusion weight $\alpha$ and evaluation threshold $\delta$ were determined experimentally.
For all of the experiments we search for the optimal fusion parameter $\alpha=\{ 0.0, 0.1, \cdots, 1.0 \}$.
Our proposed methods were implemented using the OpenCV~\cite{opencv_library} and Armadillo~\cite{Armadillo} C++ libraries.


\subsection{Short-Term Videos}

We compare the performance against two baseline systems from literature, VSUMM~\cite{deAvila2011} and VISON~\cite{Almeida2012}. 
The two baseline systems both use colour information as their primary feature.
VSUMM uses colour information by retaining only the Hue component of HSV and generating a histogram of 16 bins.
VISON is a state-of-the-art approach and consists of a histogram of the HSV representation of each frame.
It combines the HSV information in a compressed form such that the Hue component is treated with greater importance and results in a histogram of 256 bins.

An initial set of experiments were performed to find the optimal number of components for the dictionary of our proposed texture features.
Using a fixed number of components $G=\{ 8,16,32 \}$ and a fixed number of thresholds $\tau=\{ 0.05,0.10,\cdots,0.5 \}$,
we found that using just $G=8$ components provided optimal performance.
We kept the number of components constant for the remainder of our experiments.

\begin{figure}[!tb]
   \centering
   \includegraphics[width=1\columnwidth]{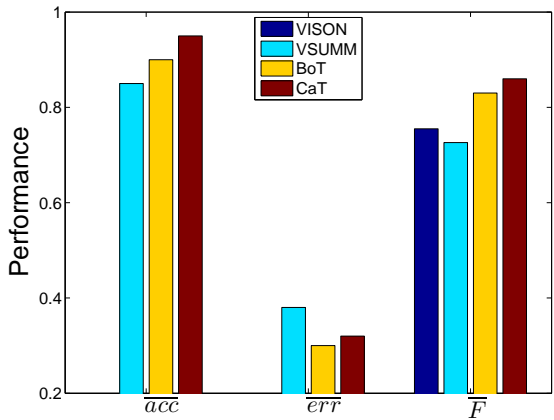}
   \caption{Comparative evaluation of our proposed methods with VSUMM~\cite{deAvila2011} and VISON~\cite{Almeida2012}.
   Lower values of $\overline{err}$ as well as higher values of $\overline{acc}$ and $\overline{F}$ are desired.}
   \label{fig:comparing}
\end{figure}

In Figure~\ref{fig:comparing} we present a summary of the average performance for 50 short-videos of our proposed systems, BoT and CaT, and the two baselines.
Two interesting results can be seen from this figure.

First, it can be seen that the texture-only BoT system performs better than either the VSUMM or VISON approaches which primarily use colour information.
The BoT system obtains an average $F$-measure of $\overline{F}=0.83$, which is a relative improvement of $9\%$ when compared to VISON, $\overline{F}=0.76$.
Furthermore, the $\overline{acc}$ and $\overline{err}$ of the BoT system shows that it produces a more accurate summarisation than VSUMM and also has the lowest $\overline{err}$ of any system%
\footnote{No results in terms of $\overline{acc}$ and $\overline{err}$ were supplied for VISON in~\cite{Almeida2012}.}.
This suggests that texture information is either equally or more important than colour information for the task of video summarisation.

Second, the proposed CaT system (fusing colour histograms and the proposed texture histograms) performs better than the two baseline systems and the proposed texture-only BoT system.
The CaT system has an average \mbox{$F$-measure} of $\overline{F}=0.86$, which is a relative improvement of $13\%$ when compared to VISON $\overline{F}=0.76$, the previous state-of-the-art approach.

\begin{figure*}
\begin{minipage}[c]{\textwidth}

	\begin{center}
		(a) ~~~~~~~~~~~~~~~~~~~~~~~
		  \includegraphics[scale=0.186]{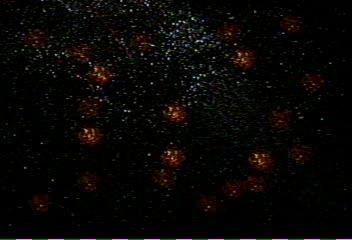}
		  \includegraphics[scale=0.186]{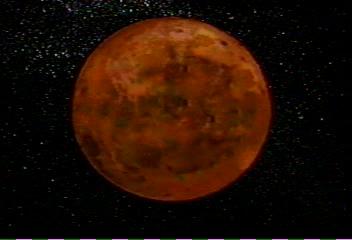}
		  \includegraphics[scale=0.186]{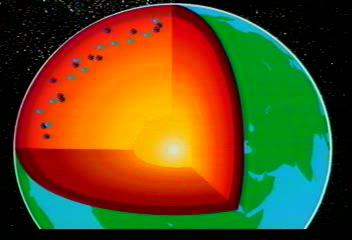}
		  \includegraphics[scale=0.186]{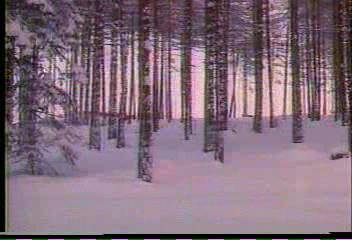}
		  \includegraphics[scale=0.186]{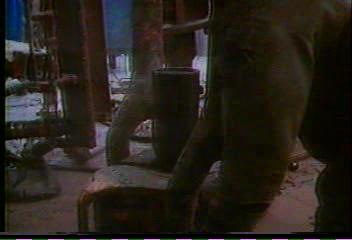}
	\end{center}

    \begin{center}

		(b) ~~    
      \includegraphics[scale=0.14]{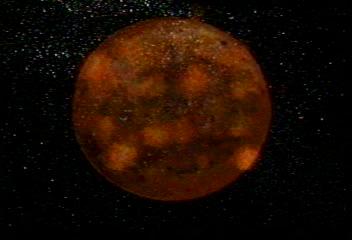}	  
	    \includegraphics[scale=0.14]{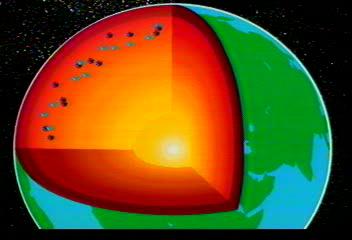}	  
	    \includegraphics[scale=0.14]{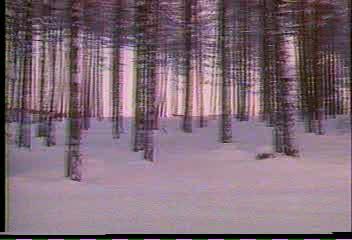}	  
	    \includegraphics[scale=0.14]{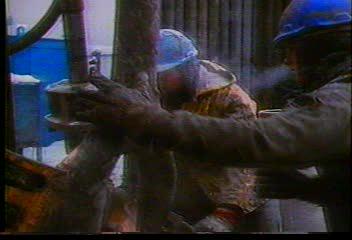}	  
	    \includegraphics[scale=0.14]{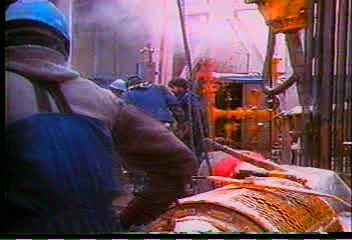}	  
	    \includegraphics[scale=0.14]{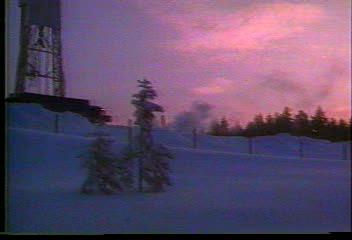}	  
	\end{center}

	\begin{center}
		(c) ~~
	  \includegraphics[scale=0.14]{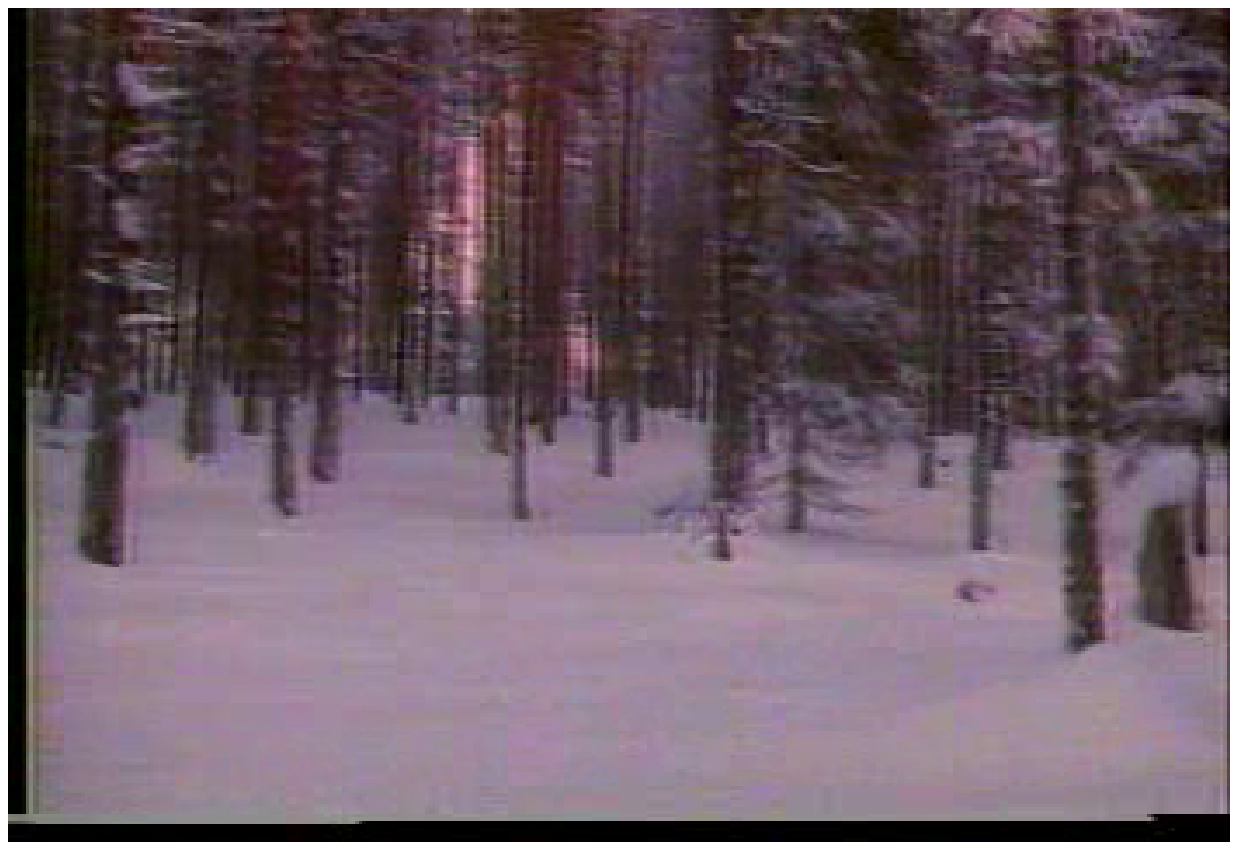}
	  \includegraphics[scale=0.14]{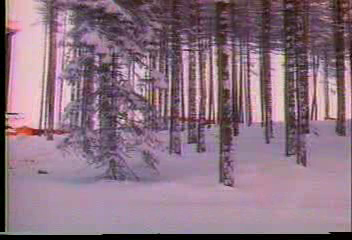}
	  \includegraphics[scale=0.14]{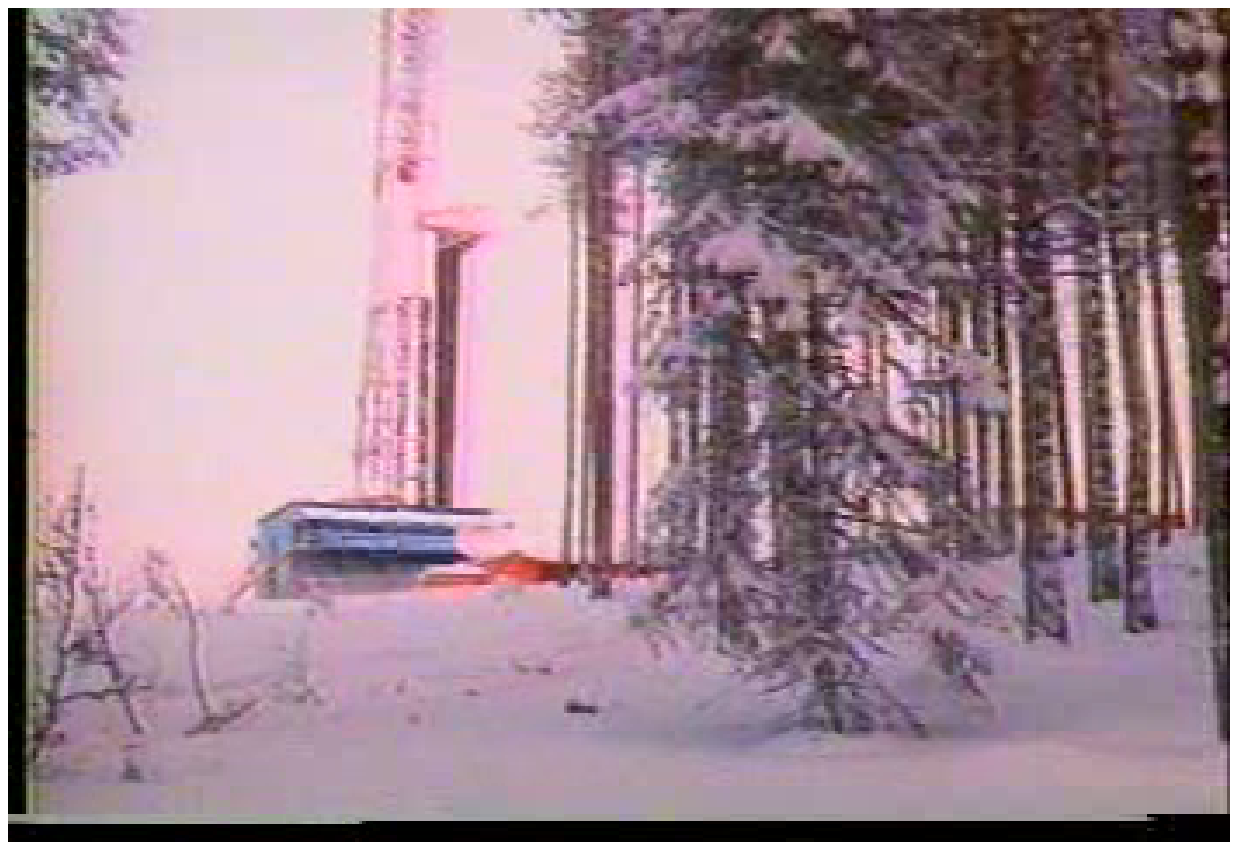}
	  \includegraphics[scale=0.14]{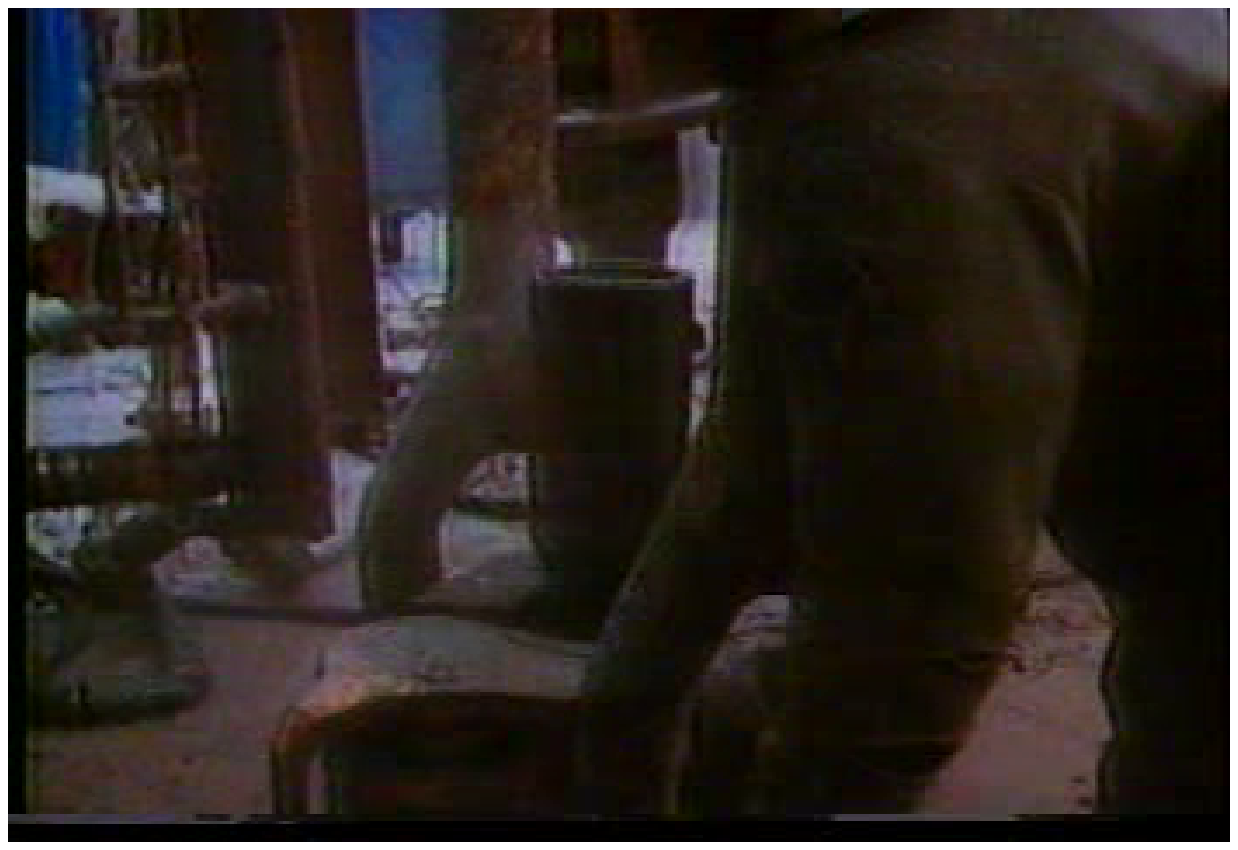}
	  \includegraphics[scale=0.14]{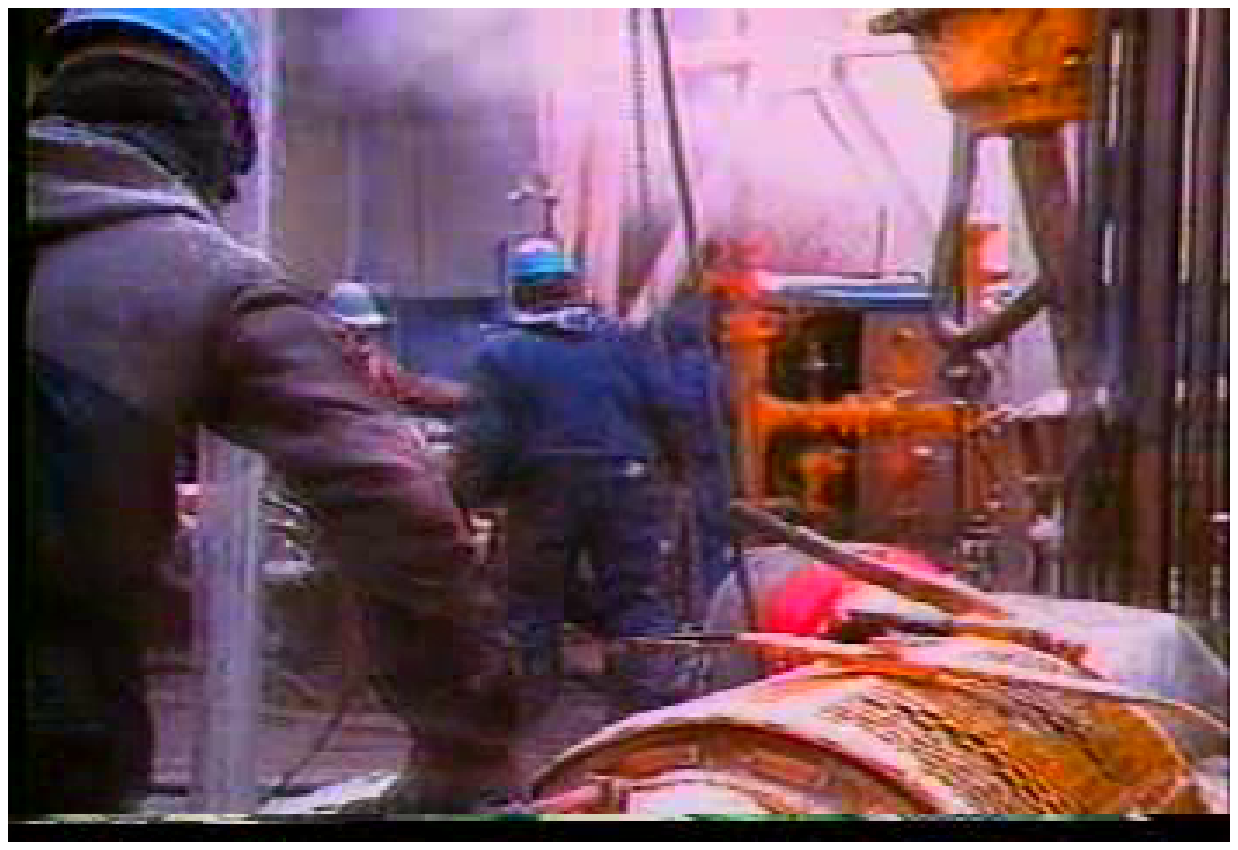}
	  \includegraphics[scale=0.14]{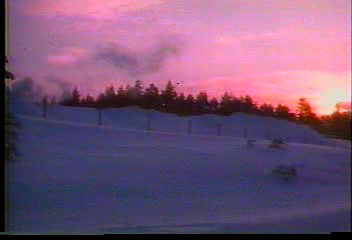}
	\end{center}
	\begin{center}
	(d) ~~~~~~~~~~~~~~~~~~~~~~~
	  \includegraphics[scale=0.14]{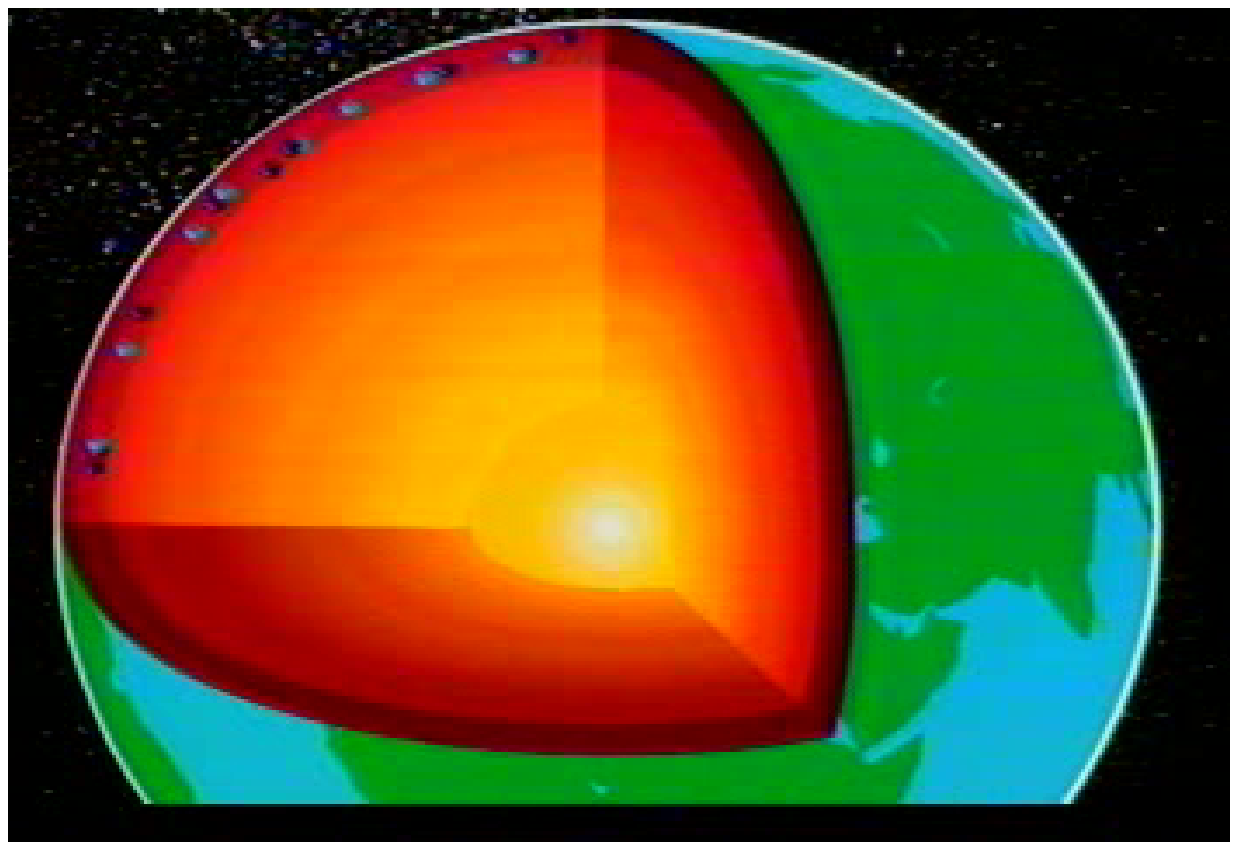}
	  \includegraphics[scale=0.14]{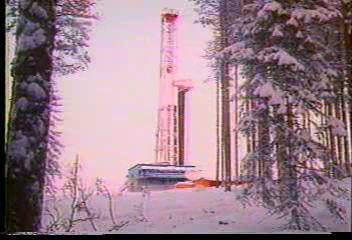}
	  \includegraphics[scale=0.14]{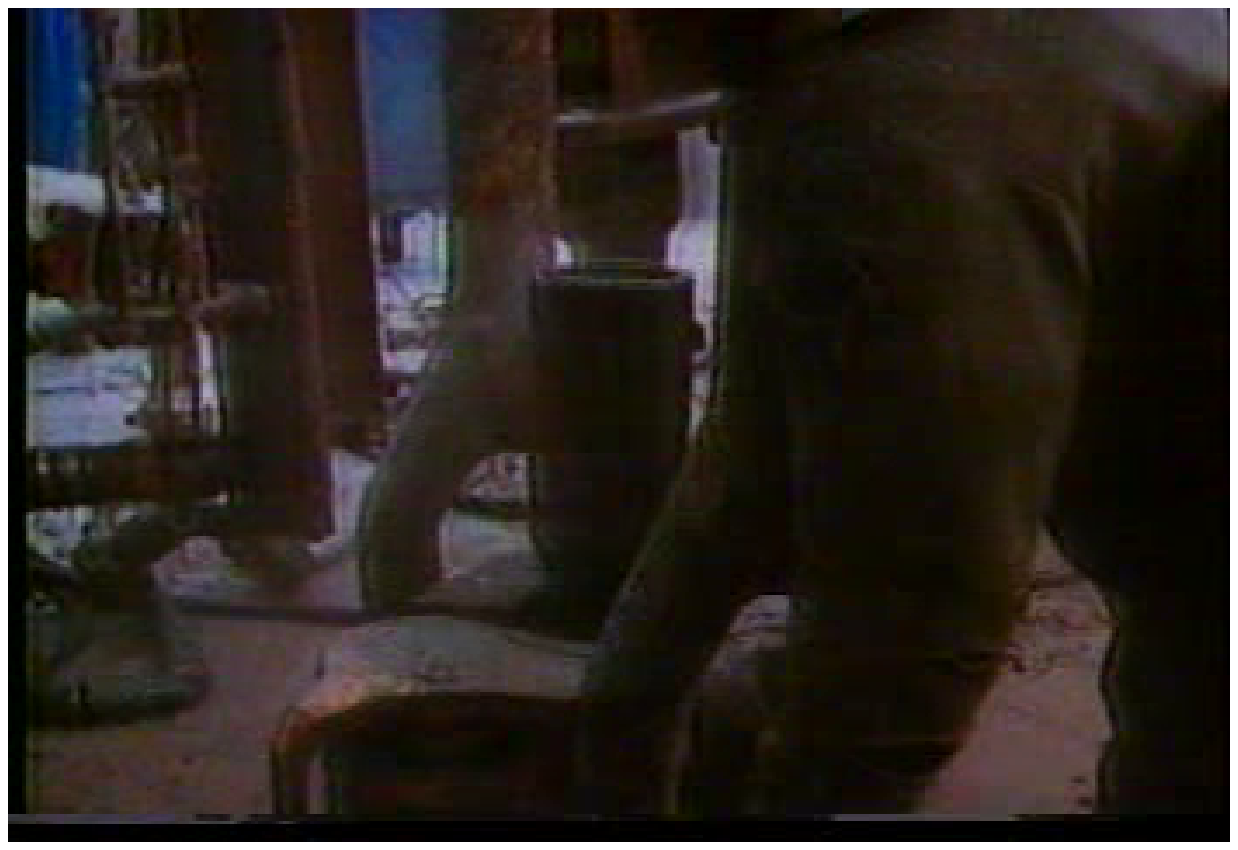}
	  \includegraphics[scale=0.14]{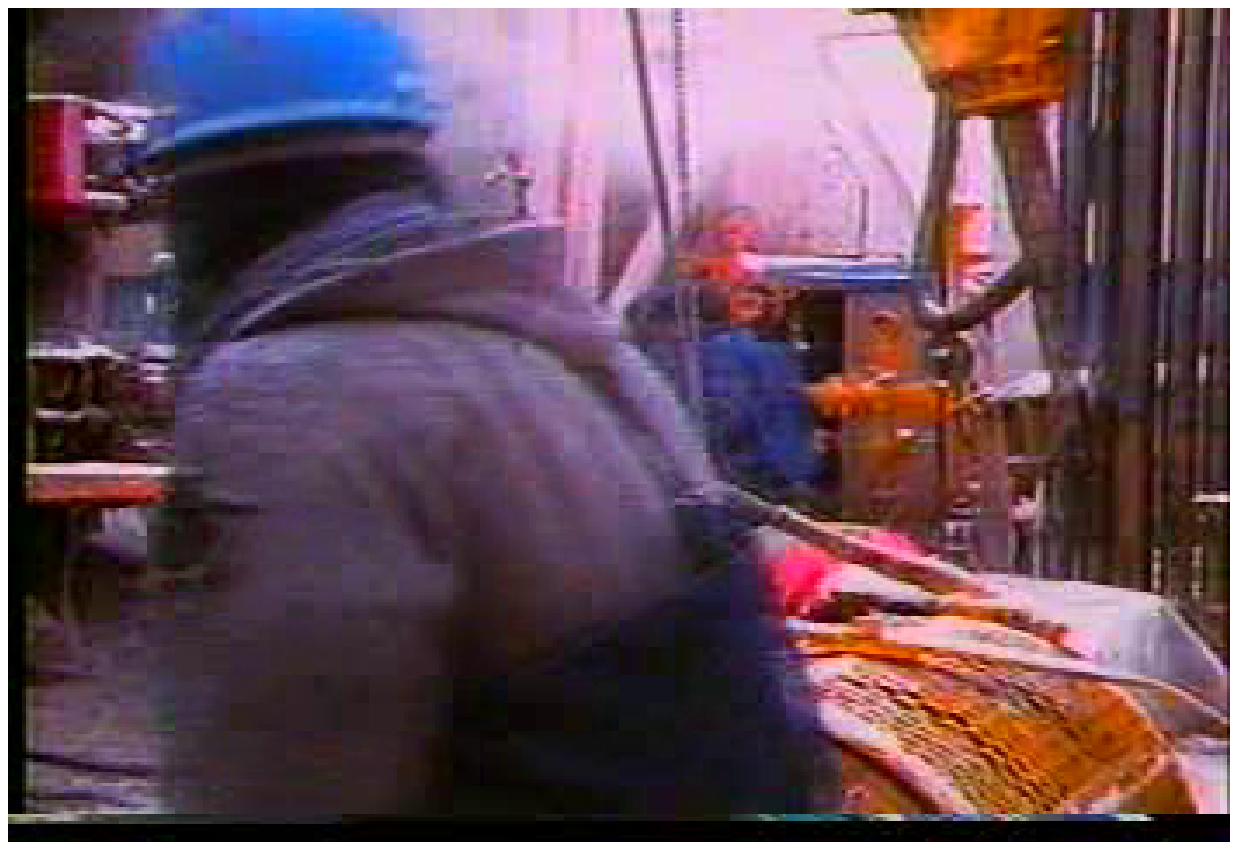}
	  \includegraphics[scale=0.14]{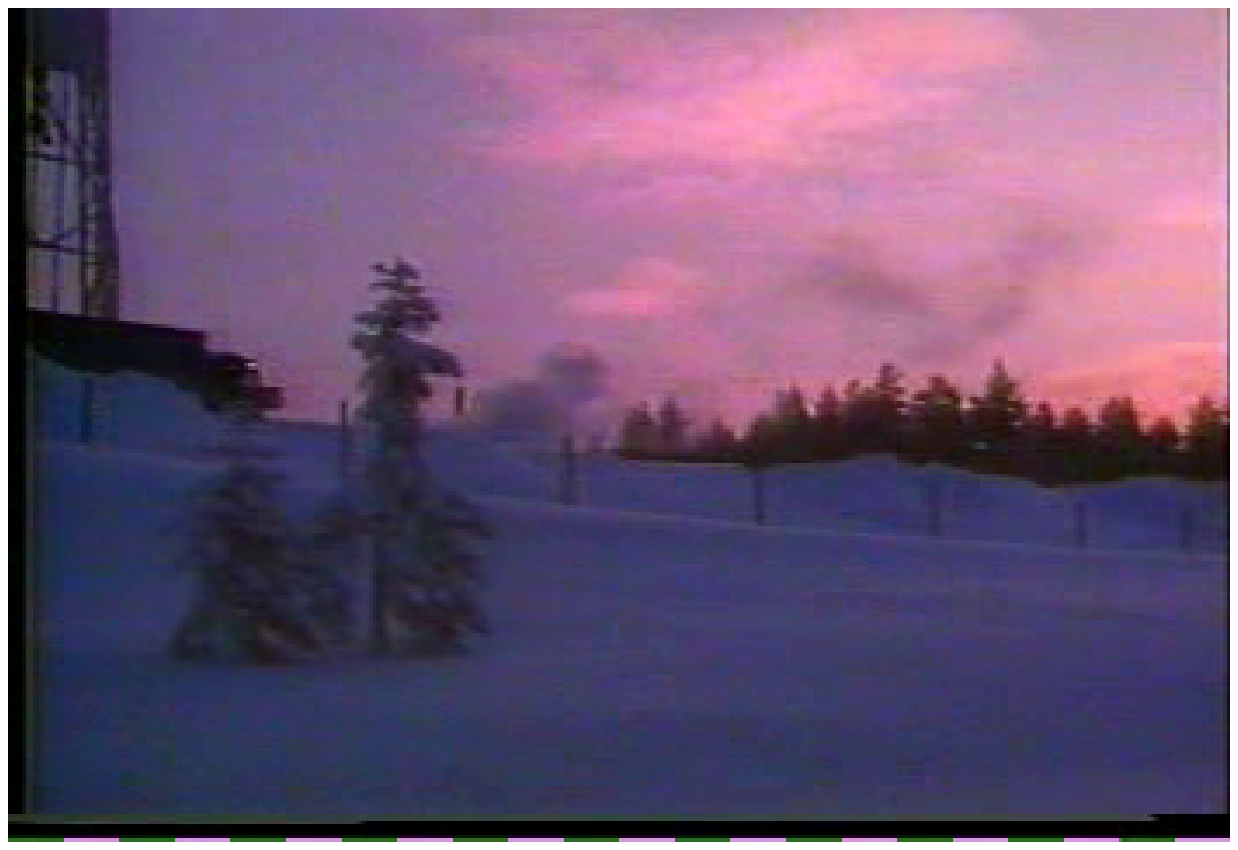}
	\end{center}   
\end{minipage}
\vspace{1ex}
\caption
  {
  Static video summary for \textit{``the future of energy gases - segment 09''},
  using 
  (a)~VSUMM,
  (b)~VISON,
  (c)~proposed BoT,
  and
  (d)~proposed CaT.
  }
\label{fig:videosummaries}
\end{figure*}

Figure~\ref{fig:videosummaries} shows the qualitative results for the automatic summarisation provided
by VSUMM and VISON as well as our proposed BoT and CaT systems.
It can be seen that VSUMM (Figure~\ref{fig:videosummaries}a) with ${F_P}=0.83$, VISON (Figure~\ref{fig:videosummaries}b) with ${F_P}=0.78$, and our proposed BoT (Figure~\ref{fig:videosummaries}c) with  ${F_P}=0.74$ contain some keyframes that may not be of interest and/or are repetitive. 
In contrast, the proposed CaT system (Figure~\ref{fig:videosummaries}d) provides the most consistent video summary with ${F_P}=0.86$. 

\subsection{Long-Term Videos}

In this section we present results on 33 long-term videos which last on average for 25 minutes. 
We examine the applicability of video summarisation to long-term videos to efficiently detect a rare species of fish and measure performance in terms of detection accuracy and compression rate (see Section~\ref{sec:eval:long_term}).

\begin{figure}[!tb]
  \centering
  \begin{minipage}{\columnwidth}
    \begin{minipage}{0.025\textwidth}
      \centering
      {\small (a)}
    \end{minipage}
    \hfill
    \begin{minipage}{0.94\textwidth}
        \centering
        \includegraphics[width=1\columnwidth]{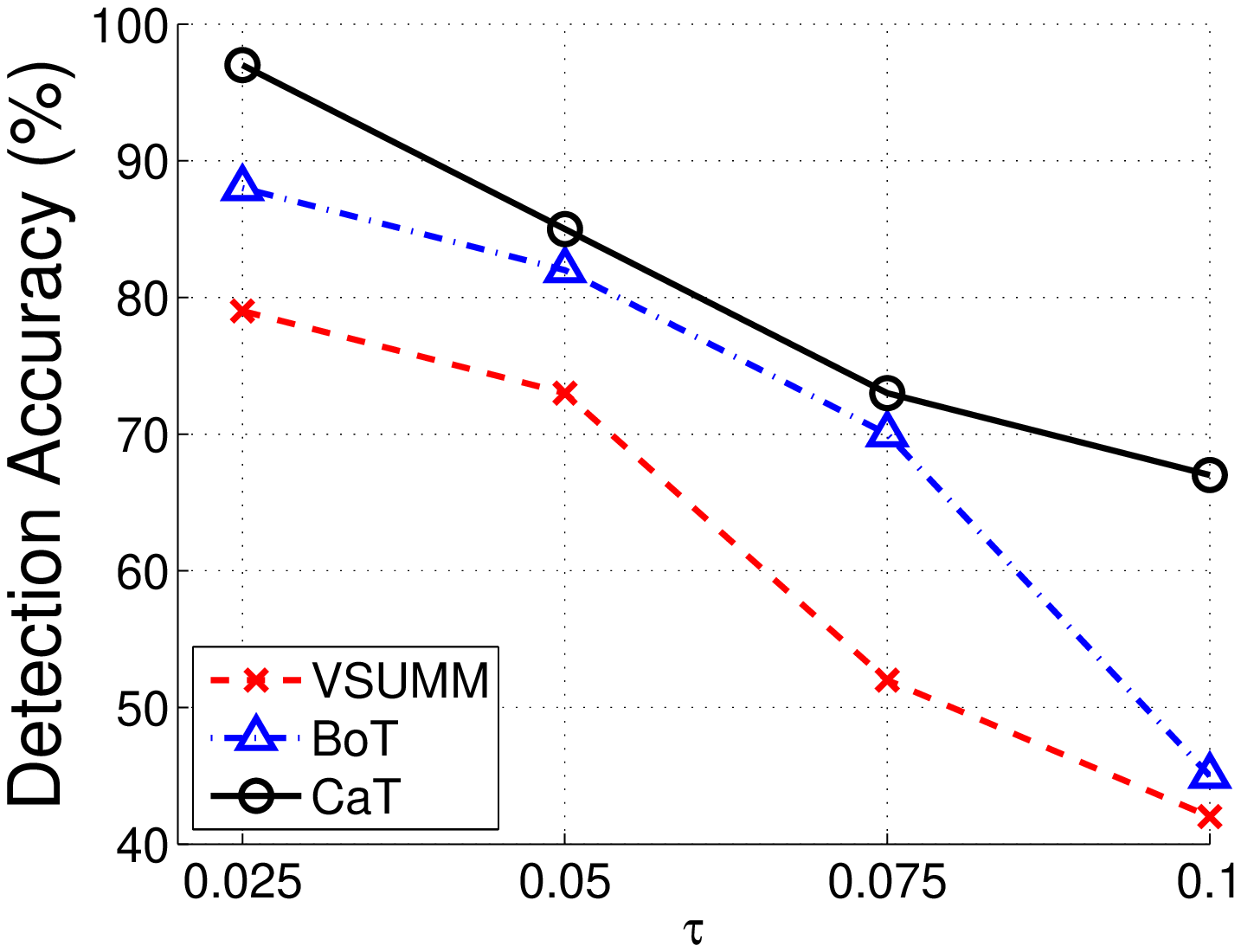}
    \end{minipage}
  \end{minipage}
  
  \begin{minipage}{\columnwidth}
    \begin{minipage}{0.025\textwidth}
      \centering
      {\small (b)}
    \end{minipage}
    \hfill
    \begin{minipage}{0.94\textwidth}
      \centering
      \includegraphics[width=1\columnwidth]{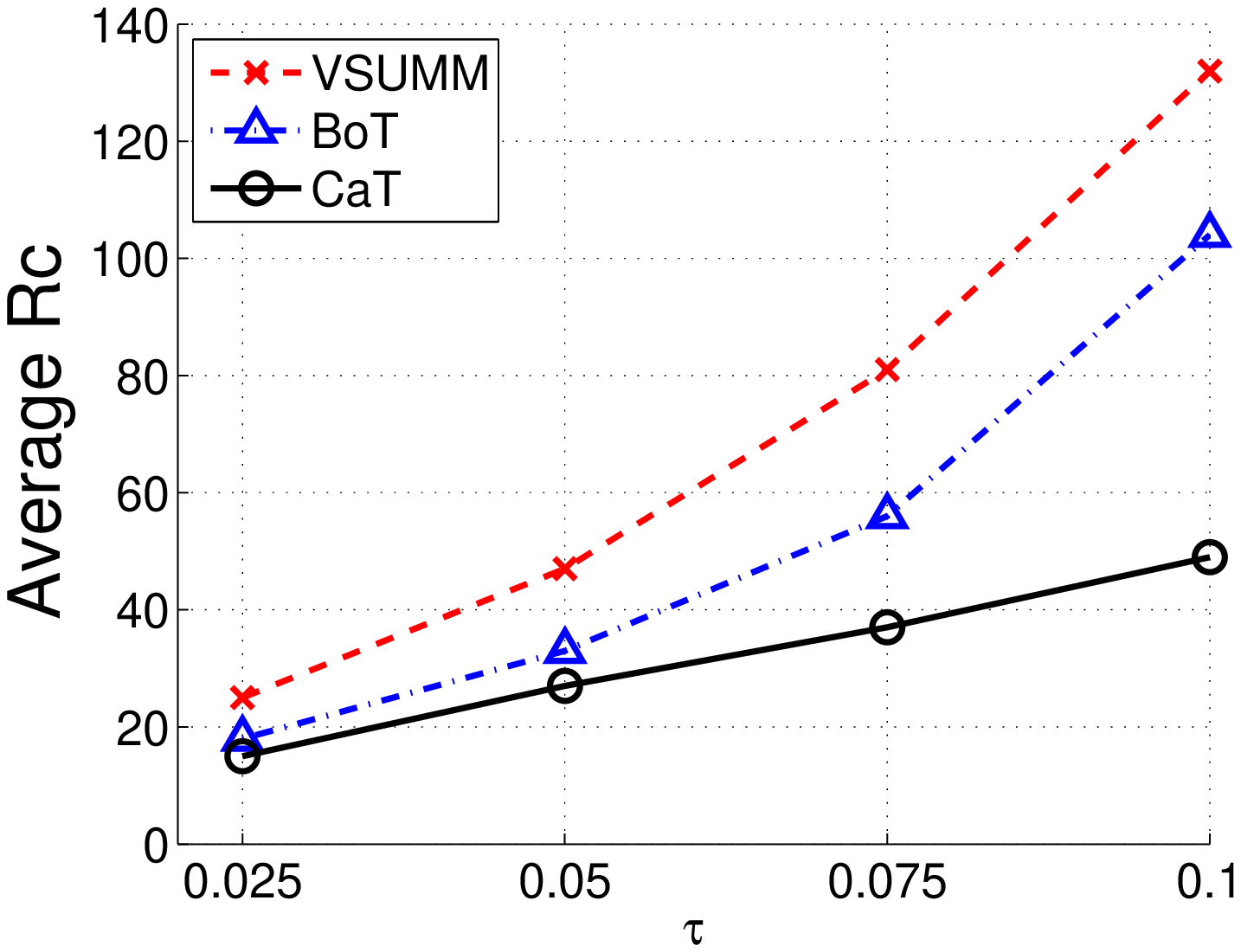}
    \end{minipage}
  \end{minipage}
  
  ~\\
  ~\\
  \caption
    {
    Demonstration of the trade-off between (a) the detection accuracy and (b) the average compression ratio $R_{c}$ for the 33 long-term videos using the CaT, BoT and VSUMM approaches.
    }
\label{fig:exp2_both}
\end{figure}

The accuracy and average compression ratio of the algorithm for various thresholds, $\tau=\{ 0.025, 0.05, \dots, 0.1 \}$, is presented in Figure~\ref{fig:exp2_both}.
It can be seen in Figure~\ref{fig:exp2_both}a that the CaT algorithm consistently outperforms the BoT and VSUMM algorithms.
We attribute this to the fact that the background in these videos is relatively stable and so the colour histograms used in VSUMM do not change as often compared to the short-term videos used in~\cite{deAvila2011}.
In Figure~\ref{fig:exp2_both}b it can be seen that while using the VSUMM algorithm provides better average compression ratio than either the BoT or CaT approaches, it comes at the cost of accuracy.
In general the proposed fusion approach provides the most consistent trade-off between accuracy and average compression ratio.

We take the optimal system at the threshold $\tau=0.05$ as this provides a high degree of detection accuracy, $85\%$, and a good average compression ratio of $27$.
This system will allow a user to see the fish of interest in $85\%$ of the summarised videos while reducing the amount of video data to view by $27$ times, more than an order of magnitude.
Such an approach would reduce the $14$ hours of video data to just $31$ minutes, thus enabling significantly more efficient reviewing of the data.

\section{Summary and Future Work}
\label{sec:conclusion}

In this paper, we have proposed the novel use of textures to perform video summarisation.
We proposed to use a visual-bag-of-textures (BoT) in two ways.
First, a BoT system which uses only texture features is proposed and it is shown to outperform two state-of-the-art systems which use colour only, VSUMM and VISON.
Second, a fused system that combines Colour and Texture (CaT) is proposed and it is shown to provide further improvements.

\newpage
Both of our proposed systems outperform two state-of-the-art approaches, VSUMM and VISON, which use colour features.
Experiments on 50 short-term videos, obtained from the Open Video Project, show that our proposed texture-only system (BoT) obtains an $F$-measure of $0.83$,
which is better than either VSUMM or VISON which obtain an average $F$-measure of $0.73$ and $0.76$, respectively.
Furthermore, our fused system (CaT) demonstrates that combining colour and texture features yields state-of-the-art performance with an average $F$-measure  of $0.86$.

We have also shown that video summarisation can be applied effectively to long-term videos.
Using 33 long-term surveillance videos, in our case underwater surveillance footage, we have shown that video summarisation can be used to significantly reduce the amount of footage to view, by up to a factor of 27, with only a minor degradation in the information content.

Future work should examine alternative features and application settings with a particular emphasis for long-term videos.
For instance, emphasising the importance of foreground objects~\cite{Reddy_TCSVT_2013} should be explored,
as well as explicit modelling of movement (or actions) of such objects~\cite{Harandi13_1,Sanin_WACV_2013}.
Moreover, the applicability of video summarisation to CCTV surveillance footage should also be considered.

\balance
{\small
\bibliographystyle{ieee}
\bibliography{BiblioJoha}
}

\end{document}